# LLMS FOR DRUG-DRUG INTERACTION PREDICTION: A COMPREHENSIVE COMPARISON

◉ **De Vito Gabriele**
Software Engineering (SeSa) Lab
University of Salerno
Salerno, Italy
gadevito@unisa.it

◉ **Ferrucci Filomena**
Software Engineering (SeSa) Lab
University of Salerno
Salerno, Italy
fferrucci@unisa.it

◉ **Athanasios Angelakis**
Department of Epidemiology and Data Science
Amsterdam UMC Locatie AMC
Amsterdam, The Netherlands
Digital Health; Methodology;
Amsterdam Public Health Research Institute
Amsterdam, The Netherlands
University of Amsterdam
Data Science Center
Amsterdam, The Netherlands
a.angelakis@amsterdamumc.nl

February 9, 2025

## ABSTRACT

The increasing volume of drug combinations in modern therapeutic regimens needs reliable methods for predicting drug-drug interactions (DDIs). While Large Language Models (LLMs) have revolutionized various domains, their potential in pharmaceutical research, particularly in DDI prediction, remains largely unexplored. This study thoroughly investigates LLMs' capabilities in predicting DDIs by uniquely processing molecular structures (SMILES), target organisms, and gene interaction data as raw text input from the latest DrugBank dataset. We evaluated 18 different LLMs, including proprietary models (GPT-4, Claude, Gemini) and open-source variants (from 1.5B to 72B parameters), first assessing their zero-shot capabilities in DDI prediction. We then fine-tuned selected models (GPT-4, Phi-3.5 2.7B, Qwen-2.5 3B, Gemma-2 9B, and Deepseek R1 distilled Qwen 1.5B) to optimize their performance. Our comprehensive evaluation framework included validation across 13 external DDI datasets, comparing against traditional approaches such as l2-regularized logistic regression. Fine-tuned LLMs demonstrated superior performance, with Phi-3.5 2.7B achieving a sensitivity of 0.978 in DDI prediction, with an accuracy of 0.919 on balanced datasets (50% positive, 50% negative cases). This result represents an improvement over both zero-shot predictions and state-of-the-art machine-learning methods used for DDI prediction. Our analysis reveals that LLMs can effectively capture complex molecular interaction patterns and cases where drug pairs target common genes, making them valuable tools for practical applications in pharmaceutical research and clinical settings.

## 1 Introduction

Drug-drug interactions (DDIs) represent a significant challenge in clinical practice, as they can alter the intended responses when patients take multiple drugs simultaneously, resulting in unexpected side effects or decreased clinical efficacy [1]. These interactions can lead to adverse drug reactions (ADRs), reduced therapeutic efficacy, or, in severe cases, life-threatening conditions [1, 2]. The risk of DDIs is particularly critical given the increasing prevalence of polypharmacy [3], where recent studies show alarming rates among older adults, ranging from 40-50% in Western countries (United States, Ireland, Sweden) to over 80% in some Asian countries (South Korea, Taiwan) [4]. With ADRs estimated to cost only the U.S. healthcare system $30.1 billion annually, and approximately 18% of these attributed to DDIs [2], predicting potential DDIs before clinical use is crucial for patient safety and successful drug development.

Traditional approaches to identifying DDIs rely on experimental methods, including in vitro and in vivo studies [5]. However, given the vast number of possible drug combinations, these approaches are time-consuming, expensive, and often impractical. Moreover, experimental testing of potentially harmful interactions raises significant ethical concerns due to the risk of adverse effects on human subjects [6].

Recent advances in computational biology and the increasing availability of molecular data have spurred the development of in silico methods for DDI prediction, aiming to complement traditional experimental approaches while addressing challenges of interpretability and clinical validation [7]. These computational approaches can be broadly categorized into three main types: literature-based extraction methods [8, 9, 10], which use natural language processes techniques to extract DDI information from biomedical literature, but are limited to documented interactions; machine learning-based prediction methods [11, 12, 13, 14], which leverage structured data from databases like DrugBank [15], but often require complex feature engineering, careful architecture design and extensive training data; and pharmacovigilance-based methods [16, 17], which can only identify DDIs after their occurrence in clinical practice [1]. Despite their success, these approaches often require careful integration of heterogeneous data sources, making them challenging to scale and adapt to new drug combinations [1].

LLMs have recently emerged as powerful tools for various biomedical tasks [18], demonstrating remarkable ability to identify hidden patterns in textual data. While LLMs have recently shown promise in various pharmaceutical applications, such as drug-target interaction prediction [19], molecule-indication translation [20], and DDI gene signature identification through knowledge graph augmentation [21], their potential for direct drug-drug interaction prediction remains largely unexplored.

This study presents the first comprehensive investigation of LLMs for DDI prediction, leveraging their ability to process multiple drug information simultaneously (SMILES notation, target organisms, and gene interactions) as text. We evaluate both zero-shot capabilities and fine-tuning approaches across 18 different LLMs, ranging from state-of-the-art models to smaller, more efficient variants. Our extensive experiments reveal several key findings. First, while LLMs demonstrate limited effectiveness in zero-shot DDI prediction (average sensitivity of 0.5463), fine-tuning significantly improves their performance. Surprisingly, smaller models like Phi-3.5 (2.7B parameters) achieve the best results, with a sensitivity of 0.978 and accuracy of 0.919, significantly outperforming both the l2-regularized logistic regression baseline [13] and larger LLMs. This performance advantage of smaller models is consistent across 13 external validation datasets, suggesting that model size is not the determining factor for DDI prediction tasks.

The main contributions of the paper are the following:

- We present the first comprehensive study of LLMs for DDI prediction, evaluating 18 models ranging from 1.5B to over 250B parameters.

- We introduce a text input representation that combines SMILES notation, target organisms, and gene interactions, enabling LLMs to leverage multiple aspects of drug information simultaneously.

- We demonstrate that while zero-shot approaches show limited effectiveness, fine-tuned smaller models can achieve superior performance compared to larger models and the l2-regularized logistic regression baseline, establishing a new state-of-the-art for DDI prediction.

- We validate our findings through extensive experimentation across 13 external datasets, confirming the robustness and generalizability of our approach.

**Structure of the paper.** The remainder of this paper is organized as follows. Section 2 provides background information on DDIs and LLMs. Section 3 reviews related work in DDI prediction. Section 4 describes our methodology, including data preparation, model selection, and evaluation framework. Section 5 presents our experimental results and analysis.

Finally, Section 6 discusses the implications of our findings and potential future directions, while Section 7 reports the conclusions.

## 2 Background

In this section, we provide essential background information to contextualize our work. We first describe drug-drug interactions, their mechanisms, and their clinical implications. We then introduce LLMs, focusing on their architecture, capabilities, and the key concepts of zero-shot learning and fine-tuning that are relevant to our study.

### 2.1 Drug-Drug Interactions

Drug-drug interactions (DDIs) occur when two or more drugs, taken simultaneously or sequentially, interact in ways that alter their individual effects. These interactions can be classified into two main categories: pharmacokinetic and pharmacodynamic interactions [22]. Pharmacokinetic interactions affect how drugs are absorbed, distributed, metabolized, or eliminated from the body, while pharmacodynamic interactions involve changes in a drug's effects at its target site. The mechanisms underlying DDIs are complex and can involve various molecular pathways. Common mechanisms include:

- Competition for drug-metabolizing enzymes (e.g., cytochrome P450).
- Alterations in drug transport proteins.
- Changes in drug absorption due to pH modifications.
- Interference with receptor binding.

Importantly, DDI prediction is a complicated problem. The order of drug administration can significantly affect the interaction outcomes, making it an inherently directional (asymmetric) problem [23, 24, 25, 26]. For instance, drug A affecting drug B's metabolism might have different implications than drug B affecting drug A's metabolism. Moreover, the same drug combination might lead to different interactions, ranging from beneficial (enhanced therapeutic effects) to adverse (increased toxicity or treatment failure).

The clinical implications of DDIs range from mild to severe. Predicting these interactions presents multiple challenges [27]. The vast number of possible drug combinations makes experimental testing impractical, while the complexity of biological pathways and the directionality of interactions add further layers of complexity. Moreover, individual patient variability in drug response and the influence of genetic factors on drug metabolism make the prediction task even more challenging. Understanding and predicting DDIs thus requires approaches capable of capturing both the molecular mechanisms of drug interactions and their directional nature. This complexity has led to the development of various computational methods, each attempting to address different aspects of the DDI prediction problem.

### 2.2 Large Language Models

LLMs are neural networks trained on vast amounts of text data to understand and generate human-like text. Through their transformer-based architecture and attention mechanisms, these models have demonstrated remarkable capabilities in various language-related tasks, from translation to summarization [28, 29, 30]. The field has seen rapid advancement with proprietary models like GPT-4 [31] and Claude 3.5 [32], alongside open-source alternatives like LLama [33], making these technologies increasingly accessible to researchers and practitioners. The development of an LLM typically involves two phases: pre-training and adaptation. Pre-training is a resource-intensive process where the model learns general language understanding from massive datasets. This foundation can then be adapted to specific tasks through different approaches. The most straightforward is zero-shot learning, where the model makes predictions without task-specific training, relying solely on its pre-trained knowledge. A more sophisticated approach is fine-tuning, where the model's parameters are adjusted using task-specific data to optimize performance for a particular application. In the context of DDI prediction, LLMs offer several compelling advantages. Their ability to process multiple drug information simultaneously allows them to handle diverse inputs, from SMILES notation (a string representation of molecular structure) to target organisms and gene interactions. Furthermore, their attention mechanisms can potentially capture complex relationships between different aspects of drug information. At the same time, their pre-training on vast amounts of biomedical literature may enable them to leverage implicit knowledge about drug interactions. However, applying LLMs to DDI prediction also presents unique challenges. The models must be carefully prompted to understand the task requirements, and their predictions must be validated against established knowledge. Moreover, the computational resources required for larger models can be substantial, making the efficiency of smaller models particularly relevant for practical applications in healthcare settings.

Given these potential benefits and challenges, understanding how LLMs can be effectively leveraged for DDI prediction represents a promising research direction. It is particularly important to explore how different model sizes and architectures affect prediction performance and identify the most efficient approaches for healthcare applications. This investigation is crucial as it could lead to more accessible and reliable tools for DDI prediction in clinical practice.

# 3 Related Work

In this section, we review existing approaches for DDI prediction, from literature-based extraction methods to machine learning and pharmacovigilance-based approaches. We conclude with recent applications of LLMs in drug discovery and the research gap our work addresses.

## 3.1 Literature-based extraction methods

Literature-based extraction methods aim to automatically identify and extract DDI information from biomedical texts, including medical reports, scientific journals, and clinical documents. These approaches typically frame DDI extraction as a relation extraction task, modeled as a multiclass classification problem. Early methods relied on conventional classifier-based approaches, particularly Support Vector Machines (SVMs), using both non-linear [34] linear kernels [8], achieving F-scores of 0.6510 in the DDIExtraction 2013 challenge [35].

More recent approaches leverage deep learning techniques, including Convolutional Neural Networks, Recurrent Neural Networks and NLP, which can automatically learn representations from text [9, 10, 36]. These models have shown superior performance to traditional approaches, with state-of-the-art methods achieving F-scores above 0.86 on standard benchmarks like the DDIExtraction 2013 corpus [35, 37].

While these methods have demonstrated success in extracting known DDIs from literature, they are inherently limited by their reliance on existing documented interactions, making them unable to predict novel, previously unreported DDIs [38].

## 3.2 Machine learning-based prediction methods

Machine learning approaches for DDI prediction can be broadly categorized into several types. Traditional approaches are based on similarity measures, where the fundamental concept is that if an interaction exists between drug A and drug B, and drug C is similar to drug A, then an interaction between drug B and drug C may occur [39]. Early works [11, 12, 40, 41] employed various similarity measures with classical algorithms such as logistic regression and SVM.

Deep learning methods (DNN) have demonstrated significant advances in this field. DNN-based approaches, such as DeepDDI [42], process structural similarity profiles through dimensionality reduction techniques before feeding them into neural networks for predicting DDI. Other approaches combine multiple drug similarities with Gaussian interaction profiles as input features [43]. Graph-based methods have emerged as powerful tools, modeling DDI prediction as a multi-relational link prediction problem on multimodal graphs incorporating drugs, proteins, and side effect relationships [44], [45]. These approaches can capture complex patterns in drug interaction networks, though they often require extensive computational resources and careful feature engineering.

Matrix factorization techniques have emerged as another effective approach for DDI prediction. For instance, ISCMF employs similarity-constrained matrix factorization on the DDI matrix, integrating eight types of similarities (including substructure, targets, and side effects) [46]. Another method, namely AMF (Adjacency Matrix Factorization) uniquely uses only known DDIs as input, sharing latent factors between rows and columns of the interaction matrix [47]. Network diffusion-based methods have also shown promise, developing an integrative label propagation framework that considers high-order similarities and feature integration [48]. In [49] a random walk-with-restart algorithm has been employed on protein-protein interaction networks to simulate signaling propagation, demonstrating how network topology can inform DDI prediction.

More recently, a more straightforward yet effective approach using drug target profiles with l2-regularized logistic regression has been proposed [13], demonstrating that gene-level information alone can achieve state-of-the-art performance. While these methods have shown promising results, they typically focus on specific types of drug information, suggesting the potential benefit of approaches capable of processing multiple drug representations simultaneously.

### 3.3 Pharmacovigilance-based methods

Pharmacovigilance-based methods focus on detecting adverse drug events induced by DDIs through the analysis of post-marketing data, playing a crucial role in public health and patient safety. These methods primarily utilize two primary data sources: Spontaneous Reporting Systems, which collect reports of suspected adverse events from healthcare professionals and patients, and Electronic Health Records, which contain both structured (e.g., laboratory results) and unstructured (e.g., clinical notes) data [50, 51]. Three main approaches characterize this field. Disproportionality analysis methods, such as those proposed in [52, 53], detect drug-adverse event combinations occurring at higher than expected frequencies. Multivariate regression approaches [54] employ logistic regression models to analyze the effects of concomitant drugs while adjusting for various factors. Association rule mining methods [55], [17], discover relationships between sets of drugs and adverse events using algorithms like Apriori. While these methods have proven valuable for post-marketing surveillance, they face several limitations. They rely heavily on reported adverse events, which may be incomplete or biased and often have significant detection latency. Moreover, they can only identify DDIs after they have occurred in clinical practice, making them less suitable for preventive screening of potential interactions.

### 3.4 LLMs in Drug Discovery

LLMs have recently emerged as promising tools in drug discovery applications. The DTI-LM, a framework leveraging LLMs for drug-target interaction prediction that processes protein amino acid sequences and drug SMILES representations has been introduced in [19]. Their approach demonstrated that while LLMs show promise in capturing protein similarities and interactions, current chemical language models still face challenges in effectively representing drug similarities. A framework leveraging protein language models (ESM-2) and chemical language models (ChemBERTa) for drug-target interaction prediction, has been introduced in [20]. Their approach demonstrated that while LLMs show promise in capturing protein similarities and interactions, current chemical language models still face challenges in effectively representing drug similarities. While showing promise, their work highlighted current limitations in chemical language models and the need for larger datasets. The DDI-GPT [21] combines LLMs with knowledge graphs for DDI prediction, achieving superior performance (AUROC 0.964) compared to existing methods and demonstrating effective zero-shot prediction capabilities. This work also provided interpretable predictions through gene importance scoring and network analysis. While these studies show the growing potential of LLMs in drug discovery, the direct application of LLMs for DDI prediction remains an emerging field with opportunities for novel research approaches.

### 3.5 Research Gap

Reviewing existing DDI prediction methodologies highlights several limitations that warrant further investigation. Literature-based extraction methods are constrained by their reliance on previously documented DDIs, inherently precluding the prediction of novel interactions [38]. While promising, machine-learning approaches often necessitate intricate feature engineering and the seamless integration of diverse, heterogeneous data sources [1]. While the efficacy of utilizing gene target information alone for prediction, it has been shown [13], more complex machine-learning methods still face these challenges. Furthermore, deep learning approaches require careful architecture design, extensive hyperparameter tuning, and large-scale training data to achieve optimal performance [1]. By their nature, pharmacovigilance-based methods are reactive, only identifying DDIs following their manifestation in clinical practice [50]. This study addresses these critical research gaps by introducing an input representation that combines multiple drug characteristics. We present a systematic evaluation of LLMs for DDI prediction, comparing their performance against a well-established baseline utilizing gene target information. This work aims to provide insights into the potential of LLMs for DDI prediction and establish a foundation for future research in this area.

## 4 Data and Method

In this section, we present our research goals, describe the datasets and data processing methods, detail our approach with LLMs (both zero-shot and fine-tuned), and outline our evaluation framework for assessing DDI prediction performance.

### 4.1 Research Goals

This study aims to investigate the effectiveness of LLMs for DDI prediction using only textual drug information (SMILES notation, target genes and organisms). We focus on three main aspects: (i) LLMs zero-shot capabilities, (ii) the impact of fine-tuning, and (iii) their performance compared to traditional approaches across multiple external datasets. More specifically, our empirical assessment is driven by three main research questions:

**RQ1** How effectively are LLMs predicting DDIs in a zero-shot setting?

This research question evaluates LLMs' ability to leverage SMILES notation, target organisms, and gene interactions to predict drug-drug interactions without task-specific training.

**RQ2** What is the impact of fine-tuning on LLMs' DDI prediction performance?

This question investigates how fine-tuning affects LLMs' ability to process multiple drug representations for DDI prediction, with particular attention to the relationship between model size and performance.

**RQ3** How do fine-tuned LLMs compare with traditional approaches?

This question assesses the performance of fine-tuned LLMs against the established l2-regularized logistic regression baseline across multiple external datasets, evaluating prediction accuracy and generalizability.

## 4.2 Datasets

Our study utilized two main data sources: DrugBank [15] and a comprehensive collection of external DDI datasets [56]. DrugBank is a comprehensive database containing detailed drug information. The dataset includes 16,581 drugs with chemical formulas (i.e., SMILES notation), target organisms, and gene interactions. DrugBank contains about 3,921 unique target genes and documents 1,420,072 known drug-drug interactions. This rich dataset is our primary source for zero-shot evaluation and fine-tuning experiments, and we also used it to generate negative examples (non-interacting drug pairs) for all datasets. For external validation, we leverage the comprehensive DDI repository [56], which aggregates drug interaction information from 14 distinct sources:

- Clinical knowledge bases: CredibleMeds, HEP, and HIV.
- Annotated corpora: DDI Corpus 2011, DDI Corpus 2013, NLM Corpus, and PK DDI Corpus.
- Healthcare systems: OSCAR EMR and WorldVista.
- Reference resources: French DDI Referrals, KEGG, and NDF-RT.
- Clinical guidelines: ONC High Priority DDI List and ONC Non-Interuptive DDI List.

Table 1 reports the initial number of drugs and DDI for each dataset.

Table 1: Drugs and Initial DDI For Datasets.

| Dataset | Drugs | DDI |
|---|---|---|
| Drugbank | 16,581 | 1,420,072 |
| CredibleMeds | 63 | 83 |
| HEP | 557 | 11,194 |
| HIV | 556 | 19,198 |
| DDI Corpus 2011 | 244 | 334 |
| DDI Corpus 2013 | 410 | 787 |
| NLM Corpus | 131 | 238 |
| PK DDI Corpus | 85 | 146 |
| OSCAR EMR | 227 | 10,325 |
| WorldVista | 378 | 13,693 |
| French DDI Referrals | 854 | 62,047 |
| KEGG | 1,033 | 52,104 |
| NDF-RT | 425 | 1,876 |
| ONC High Priority DDI List | 123 | 193 |
| ONC Non-Interuptive DDI List | 187 | 2,101 |

These diverse datasets enable a comprehensive evaluation of our approach across different contexts and data sources.

## 4.3 Data Processing

Our data processing (Figure 1) consists of four sequential steps. In the first step, we processed the DrugBank database by filtering drugs to include only those that are approved or experimental while excluding withdrawn or illicit drugs. Furthermore, we selected only drug pairs where both drugs target at least one gene, as drug target profiles are essential for our representation. For each drug pair, we extracted DrugBank IDs, SMILES notation, target organisms, and two binary vectors (length 3,921) representing gene targets, where each position corresponds to a gene in DrugBank's lexicographically ordered gene list. This processing resulted in 1,035,150 positive drug-drug interactions from DrugBank.

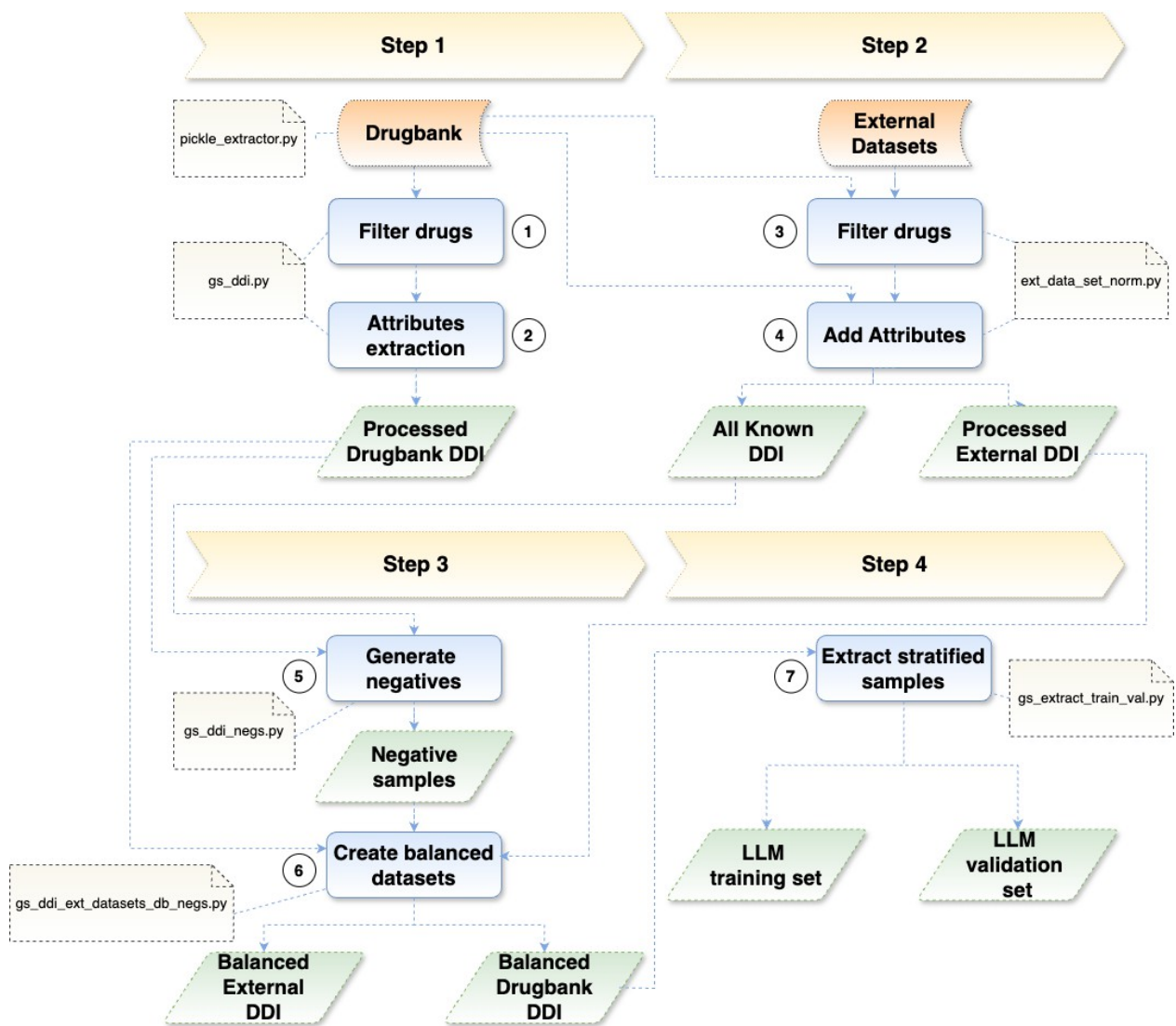

Figure 1: Data preprocessing pipeline.

In the second step, we processed the 14 datasets from [56]. We filtered out drugs not present in our processed DrugBank dataset and removed any interactions already present in our DrugBank dataset. We then enriched the remaining drug pairs with features from DrugBank (SMILES, target organisms, gene vectors). We obtained 13 usable datasets, due to the fact that the ONC High Priority DDI dataset contained no valid interactions after filtering, totaling 21,947 additional unique DDIs. In the third step, after combining all known interactions from DrugBank and external datasets (all_known_interactions), we generated 1,057,097 negative examples, ensuring no overlap with any known interaction.

We added 1,035,150 negative examples to the processed DrugBank dataset, creating a new balanced dataset. The remaining 21,947 negative examples were used to create balanced versions of the external datasets for final validation. Finally, from the DrugBank balanced dataset, we randomly extracted 1,000 examples for training and 1,090 for validation using stratified sampling. The data preprocessing pipeline and the processed datasets (with the exception of the DrugBank dataset, which cannot be publicly distributed) are available in our online repository [57] for reproducibility.

Table 2 illustrates the final distribution of drug pairs for each dataset.

## 4.4   Methods

In this section, we describe our experimental methodology. We first present the LLMs selected for our study. Then, we detail our zero-shot and fine-tuning approaches, including prompt design and training strategies.

Table 2: Datasets with Positive and Negatives Drug Pairs.

| Dataset | Drug Pairs | | |
| --- | --- | --- | --- |
| | Positive | Negative | Total |
| Drugbank | 1,035,150 | 1,035,150 | 2,070,300 |
| CredibleMeds | 5 | 5 | 10 |
| HEP | 1,271 | 1,271 | 2,542 |
| HIV | 5,651 | 5,651 | 11,302 |
| DDI Corpus 2011 | 32 | 32 | 64 |
| DDI Corpus 2013 | 74 | 74 | 148 |
| NLM Corpus | 9 | 9 | 18 |
| PK DDI Corpus | 2 | 2 | 4 |
| OSCAR EMR | 2,052 | 2,052 | 4,104 |
| WorldVista | 1,513 | 1,513 | 3,026 |
| French DDI Referrals | 4,297 | 4,297 | 8,594 |
| KEGG | 6,631 | 6,631 | 13,262 |
| NDF-RT | 119 | 119 | 238 |
| ONC High Priority DDI List | 0 | 0 | 0 |
| ONC Non-Interuptive DDI List | 291 | 291 | 582 |
| LLMs training set | 500 | 500 | 1,000 |
| LLMs validation set | 545 | 545 | 1,090 |

#### 4.4.1 Selected LLMs

Our study evaluates 18 different LLMs, ranging from 1.5B to over 250B parameters, to investigate the relationships between model size, DDI prediction performance with zero-shot and fine-tuned LLMs. These models can be categorized into four groups:

- Proprietary models: GPT-4 [31], Claude 2.1 [32], and Gemini 1.5 Pro [58], representing state-of-the-art commercial LLMs accessed through their respective APIs.

- Open-source large models: LLaMA 3.3-70B [59] and Qwen2 72B [60], which are publicly available models with architectures and parameter count comparable to commercial solutions.

- Open-source middle-range models: Granite 3.1 8B [61], LLaMa 3.1 8B [59], Gemma 2 9B [62], Falcon 3 10B [63], Mistral-Nemo 12B [64], Qwen 2.5 14B [60], Gemma2 27B [62], Aya-expanse-32b [65], and Qwen 2.5 32B [60], representing a balance between computational efficiency and model capacity.

- Open-source efficient models: Phi-3.5 (2.7B) [66], Qwen2.5 3B [60], LLaMa3.2 3B [59], and DeepSeek R1 distilled Qwen 1.5B [67], representing recent advances in efficient model architectures.

For local deployment of open-weight models, we utilized LM Studio [68], a comprehensive platform for experimenting with open-weight LLMs. LM Studio provides a unified interface for model deployment and inference, supporting various model architectures and configurations. This platform enabled us to maintain consistent experimental conditions across all open-source models while proprietary models were accessed through their respective APIs with standardized parameters. Several considerations drove the selection of these models. First, we aimed for model diversity, including large-scale and efficient architectures, to investigate the relationship between model size and performance. Second, we considered accessibility by incorporating proprietary and open-source models to assess the feasibility of DDI prediction across different deployment scenarios. Third, we included the latest models, such as Phi-3, Gemma, and DeepSeek R1, to evaluate cutting-edge architectures. Finally, we focused on smaller models (1.5B-7B parameters) to explore practical deployment options.

This comprehensive selection allows us to evaluate LLMs' general capability in DDI prediction and the specific trade-offs between model size, computational requirements, and prediction performance.

#### 4.4.2 Zero-shot Approach

To evaluate LLMs' inherent ability to predict drug-drug interactions without any task-specific training, we designed a structured prompt that incorporates all relevant drug information. Using the validation set described in Section 4.3, we formatted each drug pair into the following prompt structure:

**System prompt**

You are an expert in drug-drug interaction.
Given two drugs, where the order of administration counts, the genes and organisms targeted by the two drugs and the SMILES formulas of the two drugs, classify whether their administration causes 'interaction' or 'no interaction.' Answer only with the classification ('interaction' or 'no interaction'), nothing else.

**User prompt**

Drug1: drug1
SMILES for drug1: smiles1
Organism targeted by drug1: org1
Genes targeted by drug1: genes1
Drug2: drug2
SMILES for drug2: smiles2
Organism targeted by drug2: org2
Genes targeted by drug2: genes2
CLASSIFICATION:

This prompt design explicitly includes all available drug characteristics: drug names, molecular structures (SMILES notation), target organisms, and gene interactions. The system prompt emphasizes the importance of drug administration order and constrains the model's output to a binary classification.

For proprietary models, we submitted these prompts through their respective APIs. We leveraged LM Studio's REST API feature for open-source models, which provides an OpenAI-compatible interface for local model deployment and inference. We collected and stored the ground truth labels and the models' predictions in pickle files for subsequent analysis. This standardized approach ensures consistent evaluation across all models while maintaining the directional nature of drug-drug interactions. All evaluation scripts and corresponding results are available in our online repository [57].

### 4.4.3 Fine-tuning Strategy

Using the LLM training and validation sets described in Section 4.3, we created JSONL files containing conversational sequences structured with system prompts, user prompts (as presented in Section 4.4.2), and assistant responses. Each line in these files contains a dictionary containing each role's textual content (system, user, and assistant). For Gemma2, which does not support the system role, we concatenated the system prompt to the user prompt, resulting in JSONL files with only user and assistant interactions.

For proprietary models, we selected GPT-4 as our state-of-the-art representative for fine-tuning experiments. While Gemini also offers fine-tuning capabilities, we opted to limit our investigation to one proprietary model due to cost considerations. Claude was not included in the fine-tuning experiments as Anthropic currently does not provide fine-tuning capabilities for Claude Sonnet. For GPT-4, we utilized OpenAI's API to fine-tune the following hyperparameters: 3 epochs, batch size 3, and learning rate multiplier 0.3, which are suggested by the OpenAI documentation [69].

We selected four representatives for open-source models: Phi-3.5 2.7B, Qwen2.5 3B, Deepseek R1 Distilled Qwen 1.5B, and Gemma2 9B. This selection was motivated by investigating how smaller models perform compared to state-of-the-art models like GPT-4 and practical considerations regarding computational resources. We employed Low-Rank Adaptation (LoRA) [70] for fine-tuning these models. To optimize the fine-tuning parameters, we implemented a hyperparameter search using Optuna [71], aiming to minimize validation loss while avoiding overfitting. The search space included learning rate (log-uniform distribution in the range [1.8e-4, 2.8e-4]), number of model layers to fine-tune (16, 18, 20, 22, 24, 26, 28 for Deepseek R1 with 30 linear layers, extended to 32 for models with more layers), LoRA rank (16 or 32), LoRA alpha scaling (1 or 2, where 1 indicates alpha equals rank and 2 indicates alpha doubles rank), LoRA dropout (uniform distribution in [0.0, 0.02] with 0.001 steps), and LoRA scale (uniform distribution in [3.8, 4.4] with 0.1 steps).

We used the Adam optimizer with a cosine decay learning scheduler across 1000 trials. The optimal parameters for each model were:

- Phi-3.5 2.7B (1 and 3 epochs): layers=16, learning_rate=2e-4, rank=16, alpha=16, scale=4.0, dropout=0.0.

- Qwen2.5 3B (3 epochs): layers=16, learning_rate=2e-4, rank=16, alpha=16, scale=4.0, dropout=0.0.

- Deepseek R1 (3, 4, and 5 epochs): layers=20, learning_rate=2.2e-4, rank=32, alpha=64, scale=4.0, dropout=0.009.

- Gemma2 (5 epochs): layers=16, learning_rate=1e-5, rank=16, alpha=16, scale=4.0, dropout=0.1.

During fine-tuning, model adapters were saved every 100 iterations and merged with the base model layers upon completion. Given the non-deterministic nature of LLMs, we performed five repeated classifications on both the validation set and the external datasets described in Section 4.3 to evaluate our fine-tuned models and ensure the results' reliability. These repetitions showed no prediction variability, confirming the stability of our fine-tuned models' performance. We collected and stored the ground truth labels and the models' predictions for subsequent analysis, following the same approach used for zero-shot evaluation. All optimization scripts, fine-tuning code, and corresponding results are available in our online repository [57].

## 4.5  Evaluation Framework

To assess the performance of our approaches, we employed a comprehensive set of evaluation metrics and compared our results against an established baseline. All the experiments have been performed using a MacBook Pro M3 Max, with 96GB of RAM, 14 cores and a Metal GPU. This section details our evaluation methodology.

### 4.5.1  Metrics

We evaluated the performance of both zero-shot and fine-tuned models using several complementary metrics: 'sensitivity' measures the model's ability to correctly identify positive interactions, which is particularly crucial in drug safety applications. 'Precision' quantifies the 'accuracy' of positive predictions, while the F1-score provides their harmonic mean. We also calculated the 'accuracy' to measure general performance across both classes. Since the dataset is fully balanced regarding the 'target' (same number of positive and negative data instances), 'accuracy' is a reliable performance metric. We didn't use ROC-AUC since there are no probabilities as output from the LLMs. These metrics provide a comprehensive view of model performance, considering the critical importance of identifying dangerous interactions: 'sensitivity' and the need for reliable predictions: 'accuracy'.

### 4.5.2  Baseline Comparison

We selected the l2-regularized logistic regression model [13] as our baseline. In their original work, the authors reported impressive performance metrics, achieving an 'accuracy' of 0.9479, 'sensitivity' of 0.9556, 'specificity' of 0.948 and ROC-AUC of 0.9884. Using the balanced DrugBank dataset described in Section 4.3, we created training and validation sets (95% and 5%, respectively). Following the original methodology, we used only the drug target gene profiles as input features. The model was tuned using a stratified 10-fold cross validation exploring the regularization parameter C in the range $[2^{-16}, 2^{16}]$ specified in the original paper. We evaluated this trained model on the same LLM validation set and external balanced datasets described in Section 4.3, computing the metrics detailed in Section 4.5.1.

Our baseline comparison served two purposes: first, to try to reproduce the results of [13], and second, to provide a direct performance comparison between our LLM-based approaches and the established state-of-the-art method on the same datasets.

# 5  Results and Discussion

This section presents our experimental findings around our three research questions: zero-shot capabilities of LLMs, the impact of fine-tuning, and comparative analysis with the l2-regularized logistic regression baseline.

## 5.1  RQ1: Zero-shot Learning Analysis

Our zero-shot evaluation (see Table 3) reveals several interesting patterns across different model sizes and architectures. Proprietary models (GPT-4, Claude Sonnet, and Gemini Pro) generally demonstrated superior performance, with Claude Sonnet achieving the highest 'accuracy' (0.7358) and 'precision' (0.8859) among all models. However, even these state-of-the-art models showed relatively modest 'sensitivity' scores (0.5413-0.5927), indicating limitations in identifying positive interactions without task-specific training.

Among open-source models, we observed significant performance variations across different size categories. Some mid-range models like Gemma2 27B (accuracy: 0.7303) and Mistral-Nemo 12B (accuracy: 0.6991) performed comparably to proprietary models. However, model size did not consistently correlate with performance. For instance, some larger models like Qwen2 72B (accuracy: 0.7119) did not significantly outperform their smaller counterparts. The behavior of smaller models (2-3B parameters) was particularly interesting. While their overall accuracy was lower, some showed unusually high sensitivity scores (e.g., LLaMA-3.2 3B: 0.9982), suggesting a bias toward positive predictions rather than true discriminative ability.

Table 3: Zero-shot performance of different LLMs on DDI prediction task.

| # Params | Model | Acc. | Prec. | Sens. | F1 |
|---|---|---|---|---|---|
| >250B | Claude3.5 Sonnet | **0.7358** | 0.8859 | 0.5413 | 0.6720 |
| | Gemini 1.5 | 0.7220 | 0.7995 | 0.5927 | 0.6807 |
| | GPT-4o | 0.6459 | 0.6710 | 0.5725 | 0.6178 |
| < 2B | Deepseek Qwen1.5B | 0.4807 | 0.4677 | 0.2789 | 0.3494 |
| 2-3B | LLaMa-3.2 3B | 0.5009 | 0.5005 | **0.9982** | 0.6667 |
| | Qwen2.5 3B | 0.5037 | 0.5019 | 0.9761 | 0.6629 |
| | Phi-3.5 2.7B | 0.5358 | 0.5221 | 0.8440 | 0.6452 |
| 8-9B | Granite 3.1 8B | 0.4734 | 0.3535 | 0.0642 | 0.1087 |
| | LLaMa-3.1 8B | 0.5294 | 0.5656 | 0.2532 | 0.3498 |
| | Gemma2 9B | 0.6376 | 0.8178 | 0.3541 | 0.4942 |
| 10-14B | Falcon 3 10B | 0.5853 | 0.8394 | 0.2110 | 0.3372 |
| | Mistral-Nemo 12B | 0.6991 | 0.7395 | 0.6147 | 0.6713 |
| | Qwen 2.5 14B | 0.6569 | **0.9212** | 0.3431 | 0.5000 |
| 27-32B | Gemma2 27B | 0.7303 | 0.8476 | 0.5615 | 0.6755 |
| | Aya-Expanse 32B | 0.5917 | 0.5617 | 0.8349 | 0.6716 |
| | Qwen 2.5 32B | 0.5982 | 0.6754 | 0.3780 | 0.4847 |
| 70-72B | LLaMa-3.3 70B | 0.6477 | 0.6123 | 0.8055 | **0.6957** |
| | Qwen2 72B | 0.7119 | 0.7667 | 0.6092 | 0.6789 |

These results indicate that while LLMs can leverage their pre-trained knowledge for DDI prediction, their zero-shot performance is limited, particularly in sensitivity.

All this leads us to answer RQ1:

> **RQ₁ Zero-shot Learning Analysis:** *LLMs show limited effectiveness in zero-shot DDI prediction, with even the best models achieving only moderate accuracy and modest sensitivity. The variable performance across different model sizes suggests that pre-trained knowledge alone is insufficient for reliable DDI prediction.*

## 5.2 RQ2: Impact of Fine-tuning

Table 4 presents the performance of LLMs after fine-tuning. We selected the versions that achieved the best validation loss for open-source models experimenting with multiple epochs: Phi-3.5 trained for 3 epochs and Deepseek R1 distilled Qwen 1.5B trained for 4 epochs. For each fine-tuned model, we performed five repeated classifications to account for potential non-deterministic behavior (as described in Section 4.4.3); particularly, these repetitions showed no variability in the results, confirming the stability of our fine-tuned models.

Table 4: Fine-tuning results of different LLMs on DDI prediction task.

| Model | Acc. | Prec. | Sens. | F1 |
|---|---|---|---|---|
| GPT-4o | **0.926** | 0.922 | 0.930 | **0.926** |
| Deepseek Qwen1.5B | 0.895 | 0.877 | 0.919 | 0.898 |
| Phi-3.5 2.7B | 0.913 | 0.878 | 0.960 | 0.917 |
| Qwen2.5 3B | 0.878 | 0.820 | **0.969** | 0.888 |
| Gemma2 9B | 0.832 | **0.923** | 0.725 | 0.812 |

Fine-tuning significantly improved the performance of all models on the validation set. Most notably, smaller models showed remarkable improvements, with Phi-3.5 (2.7B parameters) achieving performance comparable to GPT-4 (accuracy: 0.913 vs 0.926) and even surpassing it in sensitivity (0.960 vs 0.930). Similarly, despite its relatively small size, Qwen2.5 3B demonstrated strong performance, particularly in sensitivity (0.969). Remarkably interesting is that even the smallest model in our study, Deepseek R1 distilled Qwen 1.5B, achieved competitive results (accuracy: 0.895, sensitivity: 0.919) after fine-tuning, suggesting that model size is not a determining factor for DDI prediction performance.

Therefore, we can answer RQ2 as follows:

> **RQ₂ Impact of Fine-tuning:** *Fine-tuning dramatically improves LLMs' DDI prediction capabilities, with even small models achieving performance comparable to or exceeding larger models. All this suggests that model size is less critical than task-specific adaptation for effective DDI prediction.*

### 5.3 RQ3: Comparative Analysis

Our replication of the l2-regularized logistic regression model achieved performance (accuracy: 0.925, sensitivity: 0.956) slightly lower than but comparable to the results reported in the original paper (accuracy: 0.948, sensitivity: 0.948).

Table 5: Sensitivity comparison across external datasets.

| Dataset | L2 (Paper) | L2 (Repl.) | Phi3.5 3b | Qwen2.5 3B | Gemma2 9B | GPT-4o +250B | Deepseek Qwen 1.5B |
|---|---|---|---|---|---|---|---|
| | | | | Sensitivity | | | |
| CredibleMeds | 0.800 | **1.000** | **1.000** | **1.000** | 0.400 | **1.000** | **1.000** |
| HEP | 0.974 | 0.971 | **1.000** | **1.000** | 0.725 | 0.917 | 0.990 |
| HIV | 0.900 | 0.920 | **1.000** | **1.000** | 0.765 | 0.982 | 0.971 |
| Corpus 2011 | **1.000** | 0.938 | 0.938 | 0.938 | 0.906 | 0.750 | 0.906 |
| Corpus 2013 | 0.774 | 0.905 | **1.000** | 0.986 | 0.905 | 0.905 | 0.905 |
| NLM Corpus | 0.800 | 0.889 | 0.889 | 0.889 | 0.889 | 0.889 | **1.000** |
| PK Corpus | **1.000** | **1.000** | **1.000** | **1.000** | **1.000** | **1.000** | **1.000** |
| OSCAR | 0.899 | 0.866 | **0.942** | 0.936 | 0.696 | 0.905 | 0.872 |
| WorldVista | 0.924 | 0.907 | **0.986** | 0.984 | 0.851 | 0.841 | 0.970 |
| French Ref. | 0.883 | 0.928 | 0.972 | **0.986** | 0.808 | 0.981 | 0.879 |
| KEGG | 0.950 | 0.891 | **0.990** | 0.985 | 0.714 | 0.962 | 0.956 |
| NDF-RT | 0.970 | 0.975 | **1.000** | 0.966 | 0.773 | **1.000** | 0.966 |
| Onc Non-Int. | 0.875 | 0.835 | **1.000** | **1.000** | 0.904 | 0.993 | 0.969 |
| **AVG** | 0.904 | 0.925 | **0.978** | 0.975 | 0.787 | 0.933 | 0.953 |

Table 6: Accuracy comparison across external datasets.

| Dataset | L2 (Repl.) | Phi3.5 3b | Qwen2.5 3B | Gemma2 9B | GPT-4o +250B | Deepseek Qwen 1.5B |
|---|---|---|---|---|---|---|
| | | | Accuracy | | | |
| CredibleMeds | **1.000** | **1.000** | 0.900 | 0.700 | **1.000** | **1.000** |
| HEP | 0.919 | **0.924** | 0.895 | 0.823 | 0.908 | 0.923 |
| HIV | 0.903 | 0.927 | 0.898 | 0.846 | **0.948** | 0.915 |
| Corpus 2011 | 0.922 | 0.875 | 0.859 | **0.938** | 0.797 | 0.844 |
| Corpus 2013 | 0.899 | **0.919** | 0.878 | 0.872 | 0.878 | 0.865 |
| NLM Corpus | 0.889 | 0.778 | 0.667 | **0.944** | 0.833 | 0.889 |
| PK Corpus | **1.000** | 0.500 | 0.750 | 0.750 | 0.750 | 0.750 |
| OSCAR | 0.882 | 0.902 | 0.872 | 0.821 | **0.913** | 0.874 |
| WorldVista | 0.905 | **0.924** | 0.899 | 0.892 | 0.877 | 0.919 |
| French Ref. | 0.910 | 0.910 | 0.893 | 0.868 | **0.947** | 0.869 |
| KEGG | 0.889 | 0.922 | 0.895 | 0.824 | **0.937** | 0.910 |
| NDF-RT | 0.971 | 0.958 | 0.908 | 0.878 | **0.983** | 0.950 |
| Onc Non-Int. | 0.852 | 0.919 | 0.897 | 0.907 | **0.952** | 0.909 |
| **AVG** | **0.918** | 0.881 | 0.862 | 0.851 | 0.902 | 0.894 |

As reported in Section 5.2, on the validation set, our fine-tuned LLMs showed competitive performance, with GPT-4 (accuracy: 0.926, sensitivity: 0.930) and Phi-3.5 (accuracy: 0.913, sensitivity: 0.960) achieving comparable results to our replicated baseline. The external validation across 13 datasets provided comprehensive insights. Regarding sensitivity scores (see Table 5), while the original paper reported an average of 0.904, our fine-tuned smaller models demonstrated exceptional performance, with Phi-3.5 and Qwen2.5-3B achieving an average of 0.978 and 0.953, surpassing both GPT-4 (0.933) and the baseline results. For 'accuracy' (Table 6), our replicated baseline achieved strong performance (average: 0.918), with GPT-4 showing comparable results (0.902) and Phi-3.5 maintaining competitive performance (0.881). Similarly, for F1-scores (Table 7), the replicated baseline (average: 0.919) was closely matched by GPT-4 (0.906) and Phi-3.5 (0.900).

Table 7: F1 comparison across external datasets.

| Dataset | F1-score | | | | | |
| | L2 (Repl.) | Phi3.5 3b | Qwen2.5 3B | Gemma2 9B | GPT-4o +250B | Deepseek Qwen 1.5B |
|---|---|---|---|---|---|---|
| CredibleMeds | **1.000** | **1.000** | 0.909 | 0.571 | **1.000** | **1.000** |
| HEP | 0.923 | **0.930** | 0.905 | 0.804 | 0.908 | 0.928 |
| HIV | 0.904 | 0.932 | 0.907 | 0.833 | **0.950** | 0.919 |
| Corpus 2011 | 0.923 | 0.882 | 0.870 | **0.935** | 0.787 | 0.853 |
| Corpus 2013 | 0.899 | **0.925** | 0.890 | 0.861 | 0.882 | 0.870 |
| NLM Corpus | 0.889 | 0.800 | 0.727 | **0.941** | 0.842 | 0.900 |
| PK Corpus | **1.000** | 0.667 | 0.800 | 0.800 | 0.800 | 0.800 |
| OSCAR | 0.880 | 0.905 | 0.880 | 0.795 | **0.913** | 0.874 |
| WorldVista | 0.905 | **0.928** | 0.907 | 0.887 | 0.873 | 0.923 |
| French Ref. | 0.912 | 0.915 | 0.902 | 0.860 | **0.949** | 0.871 |
| KEGG | 0.889 | 0.927 | 0.903 | 0.803 | **0.939** | 0.914 |
| NDF-RT | 0.971 | 0.960 | 0.913 | 0.864 | **0.983** | 0.950 |
| Onc Non-Int. | 0.850 | 0.925 | 0.907 | 0.907 | **0.954** | 0.914 |
| AVG | **0.919** | 0.900 | 0.878 | 0.836 | 0.906 | 0.901 |

Given the results, we can answer RQ3 as follows:

> **RQ3 Comparative Analysis:** *Fine-tuned LLMs demonstrate performance comparable to the l2-regularized logistic regression on both the validation set and external datasets. Notably, smaller models like Phi-3.5 and Qwen2.5-3B achieve superior sensitivity scores while maintaining competitive accuracy and F1-scores.*

## 6 Implications, Limitations and Future Work

This section discusses the practical implications of our findings and outlines limitations and future research directions.

### 6.1 Practical Implications

Our findings have significant implications for both the deployment of DDI prediction systems and their clinical applications. From a deployment perspective, the superior performance of smaller models (2-3B parameters) represents a breakthrough in accessibility. These models can run efficiently on standard computing hardware, making them viable for implementation in various healthcare settings without requiring specialized infrastructure. The ability to deploy locally also addresses critical privacy concerns, as sensitive medical data can be processed on-premises rather than through external APIs. Moreover, the lightweight nature of these models enables potential integration into existing healthcare information systems, including electronic health records.

From a clinical perspective, our results suggest several promising applications. The high sensitivity achieved by our models, particularly in detecting known interactions across diverse external datasets, indicates their potential as reliable screening tools in clinical practice. This potential is especially valuable for complex polypharmacy cases, where traditional approaches might miss potential interactions. The models' ability to process multiple drug representations simultaneously (SMILES notation, target organisms, and gene interactions) could provide more comprehensive interaction assessments than current single-feature approaches.

Furthermore, these models could support clinical decision-making at various levels:

- Pre-prescription screening to identify potential interactions before medication is prescribed
- Support for pharmacists in medication review processes
- Aid in clinical research for identifying potential drug combinations for investigation
- Real-time decision support in emergency medicine where rapid assessment of drug interactions is crucial

The combination of computational efficiency and robust performance suggests these models could serve as practical tools in everyday clinical workflows, potentially improving patient safety while maintaining operational efficiency.

## 6.2 Future Directions

While our study demonstrates the potential of LLMs for DDI prediction, several promising research directions warrant further investigation. The strong performance of smaller models suggests exploring even more efficient architectures specifically designed for molecular and biological data processing, for instance, developing specialized pre-training strategies that incorporate domain-specific knowledge about drug interactions and biological pathways. Another important direction is the integration of additional drug-related information. While our current approach combines SMILES notation, target organisms, and gene interactions, future work could incorporate other relevant data such as pharmacokinetic properties, metabolic pathways, and temporal aspects of drug administration. Integrating additional information could lead to more nuanced predictions about the timing and severity of potential interactions. The interpretability of model predictions represents another crucial area for future research. Developing methods to explain why specific drug combinations are flagged as potentially dangerous would increase trust in these systems and provide valuable insights for healthcare professionals, involving techniques for analyzing attention patterns or developing attribution methods specific to drug interaction prediction. Finally, investigating the models' ability to handle novel drug compounds and rare interactions could enhance their practical utility. This exploration might implicate few-shot learning approaches or techniques for continuous model updating as new drug interaction data becomes available. Additionally, exploring the models' potential for predicting not just the presence of interactions but also their mechanisms and severity levels could provide more comprehensive support for clinical decision-making.

## 6.3 Threats to Validity

Several factors could potentially threaten the validity of our study. Regarding internal validity, our experimental setup and controls required careful consideration. The selection of fine-tuning parameters might have affected model performance; we mitigated this through systematic hyperparameter optimization using Optuna. The random split between training and validation sets could have also impacted results; we addressed this by validating our findings across 13 external datasets. Following the methodology in [13], we used balanced datasets to prevent learning bias.

Concerning external validity and the generalizability of our findings to future scenarios, while our fine-tuning experiments were limited to five selected models (one proprietary and four open-source), we mitigated this threat by choosing models with different architectures and sizes (from 1.5B to over 250B parameters) and validating our approach across 13 diverse external datasets.

To ensure conclusion validity and the reliability of our results, we performed five repeated classifications for fine-tuned models (RQ2 and RQ3), which showed no variability in the results.

# 7 Conclusion

This study represents the first comprehensive investigation of LLMs for drug-drug interaction prediction. Our findings demonstrate that while LLMs show limited effectiveness in zero-shot settings, fine-tuned models achieve remarkable performance, with smaller models like Phi-3.5 (2.7B parameters) performing comparably to or better than larger models and traditional approaches. The good performance of small LLMs has significant practical implications, making DDI prediction more accessible for clinical applications without requiring extensive computational resources. Our validation across 13 external datasets confirms the robustness and generalizability of the approach. These results establish LLMs as powerful tools for DDI prediction and suggest that model size is less critical than task-specific adaptation for effective performance. Future work could focus on developing even more efficient architectures, incorporate additional drug-related information, and improve model interpretability for clinical applications.